\pdfoutput=1

\documentclass[11pt]{article}

\usepackage{EMNLP2023}

\usepackage{times}
\usepackage{latexsym}

\usepackage[T1]{fontenc}

\usepackage[utf8]{inputenc}

\usepackage{microtype}

\usepackage{inconsolata}

\usepackage{times}
\usepackage{latexsym}
\usepackage{enumitem}
\usepackage{xcolor}
\usepackage{amsmath}
\usepackage{graphicx}
\usepackage{amsfonts}
\usepackage{comment}
\usepackage{booktabs}
\usepackage{url}
\usepackage{multirow}
\usepackage{subcaption}

\newcommand{\x}{\boldsymbol{x}}
\newcommand{\y}{\boldsymbol{y}}
\newcommand{\model}{Self-Cleaning}

\title{Improving a Named Entity Recognizer Trained on Noisy Data\\ with a Few Clean Instances}

\author{
    Zhendong Chu\textsuperscript{1},
    Ruiyi Zhang\textsuperscript{2},
    Tong Yu\textsuperscript{2},
    Rajiv Jain\textsuperscript{2}, \\
    \textbf{Vlad I Morariu\textsuperscript{2},
     Jiuxiang Gu\textsuperscript{2},
    Ani Nenkova\textsuperscript{2}}
    \\ 
    \textsuperscript{1}University of Virginia, Charlottesville, VA, USA \\
    \textsuperscript{2}Adobe Research, San Jose, CA, USA \\
    \texttt{zc9uy@virginia.edu},
    \texttt{\{ruizhang,tyu,rajijain,morariu,jigu,nenkova\}@adobe.com}
}

\begin{document}
\maketitle
\begin{abstract} 
To achieve state-of-the-art performance, one still needs to train NER models on large-scale, high-quality annotated data, an asset that is both costly and time-intensive to accumulate. In contrast, real-world applications often resort to massive low-quality labeled data through non-expert annotators via crowdsourcing and external knowledge bases via distant supervision as a cost-effective alternative. However, these annotation methods result in noisy labels, which in turn lead to a notable decline in performance. Hence, we propose to denoise the noisy NER data with guidance from a small set of clean instances.  Along with the main NER model we  train a discriminator model and use its outputs to recalibrate the sample weights. The discriminator is capable of detecting both span and category errors with different discriminative prompts. Results on public crowdsourcing and distant supervision datasets show that the proposed method can consistently improve performance with a small guidance set.
\end{abstract}

\section{Introduction}
Deep learning methods have notably improved the performance of named entity recognition (NER), but need large-scale high-quality labeled data \citep{lample2016neural, devlin2018bert}.
In practice, collecting large-scale labeled data via crowdsourcing \citep{rodrigues2018deep, finin2010annotating} or distant supervision \citep{liang2020bond} is far more cost-effective. However, such data is usually too noisy for direct use  without further treatment \cite{hedderich2021analysing, liang2020bond, chu2020insrl}. 

Extensive efforts have been dedicated to develop data denoising techniques and learning strategies specifically tailored for noisy NER data. \citet{liang2020bond} suggested fine-tuning pre-trained language models (PLMs) on such data, employing early stopping and self-training techniques to mitigate overfitting induced by noisy labels. \citet{meng2021distantly} extended the approach by using a frozen PLM to generate augmented pseudo labels for self-training. \citet{liu2021noisy} further eliminated self-training labels with low estimated label confidence.
Yet these denoising methods do not have a mechanism to  guide error correction, thus suffer from \emph{confirmation bias} \citep{tarvainen2017mean, arazo2020pseudo}, where the learner struggles to correct its own mistakes.

\begin{figure}[htp!]
    \centering
    \includegraphics[width=7.5cm]{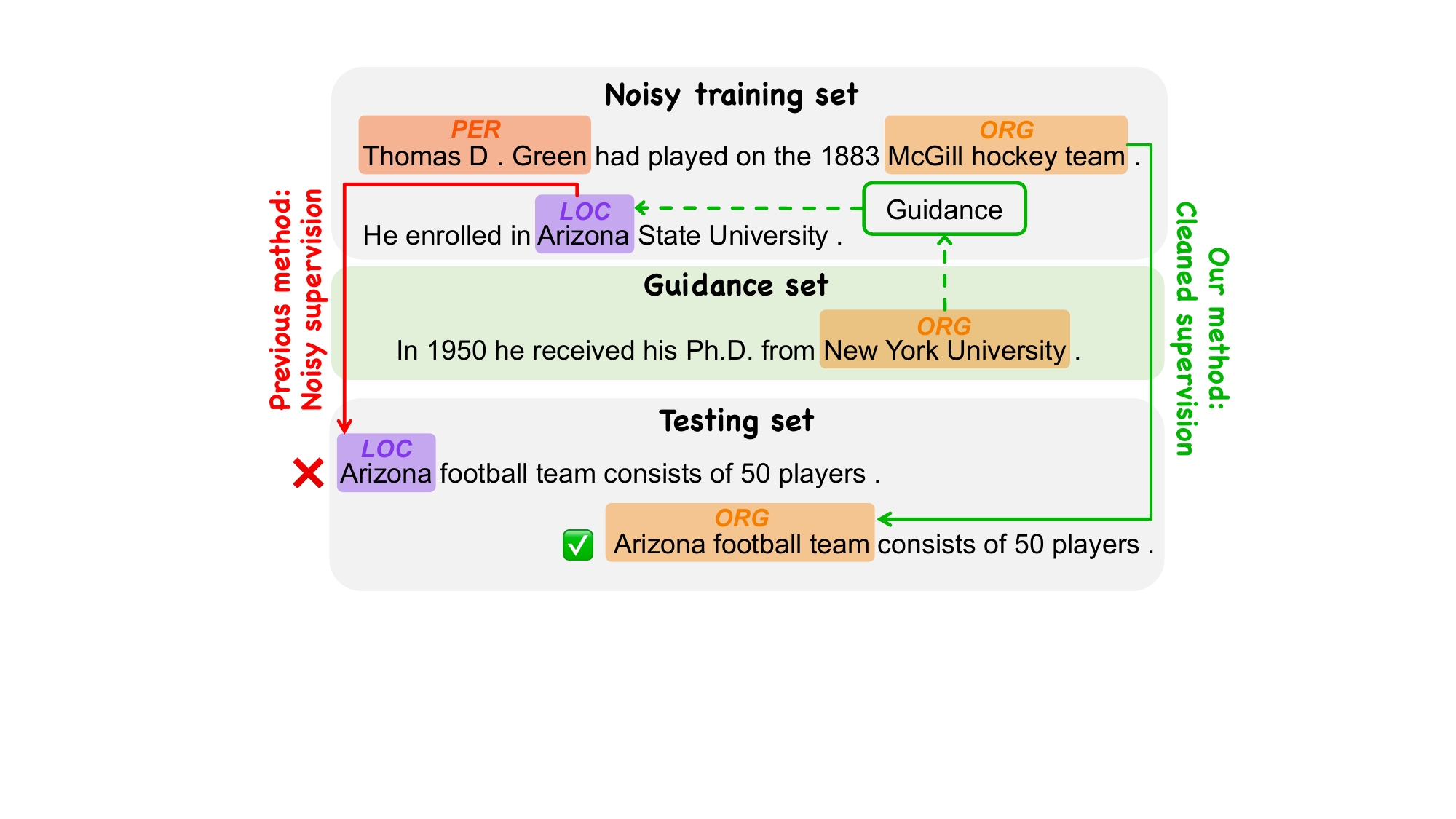}
    \caption{Illustration of the Guided Denoising Framework. The initial noisy label, \texttt{Arizona-LOC}, presents a deviation from the patterns observed in the guidance set, where geographical names preceding the term \texttt{University} are appropriately categorized into an organization entity (e.g., \texttt{New York University-ORG}). The depicted process of guided denoising (highlighted in green) ensures the retention of the accurately supervised label, \texttt{McGill hockey team-ORG}, thereby facilitating the acquisition of correct entity recognition patterns.}
    \label{fig:example}
    \vspace{-4mm}
\end{figure}

One natural idea to improve the performance of NER models trained on noisy data is to incorporate a small set of clean instances, which can be obtained at an acceptable cost. For example, one can let a financial professional manually label a subset of financial named entities and use them to guide the learning process on a larger, distantly-supervised financial NER dataset. We refer to the small clean set as the \emph{guidance} set. There are a number of possibilities of how to effectively utilize the guidance set. The most straightforward method is to further fine-tune the model trained on the noisy NER data on the guidance set; we treat this approach as a baseline to compare against. 
\citet{jiang2021named} augmented the noisy labels with a confidence score according to their probability of being correct given the clean data. Their heuristic-based approach is not tailored to the noisy NER problem, and as a result, it fails to identify particular types of noise in NER, such as span errors. We present a complementary approach that is effective in correcting NER-specific errors.

We propose a \emph{Guided Denoising Framework} (shown in Figure \ref{fig:example}) to better utilize the guidance data by eliminating noisy labels that conflict with the patterns in the guidance set.
In this framework, in addition to the NER model, we also use a discriminator specifically designed to detect such conflicts.
 This discriminator is responsible for evaluating the accuracy of assigned labels and is trained in a few-shot manner \citep{brown2020language, liu2021pre} with the small guidance set. 
 Based on the analysis of real-world noisy NER datasets, we equip the discriminator, by designing different prompts, with the ability to detect two error types: span error and category error. The outputs of the discriminator are used to reweight the samples for the NER model's training. We also design a co-training strategy to improve the discriminator and the NER model in a collaborative manner. In short, we make the following contributions:
 
\begin{itemize}[leftmargin=*]
    \item We propose \textbf{Self-Cleaning}, a generic guided denoising framework for improving NER learning on noisy data with a small guidance set. To the best of our knowledge, this is the first instance of a denoising framework making use of an auxiliary model to correct noise in the data. 
    \item We design a prompt-based discriminator to detect noisy NER labels. The discriminator is capable of identifying both span errors and category errors in the noisy NER data using distinct prompts.
    \item We report extensive experiments and ablation studies on NER benchmarks with crowdsourcing and distant supervision NER data. Results show that our approach boosts the performance. 
\end{itemize}

\section{Background}
\subsection{Named Entity Recognition}
NER is the task of identifying named entities in plain text and classifying them into pre-defined entity categories, such as person, organizations, locations, etc \citep{li2020survey}. Formally, we denote a sentence consisting of $n$ tokens as $\boldsymbol{x}=[x_1, ..., x_n]$  and their corresponding labels as $\boldsymbol{y}=[y_1, ..., y_n]$. We define $D=\{(\x_i, \y_i)\}_{i=1}^{|D|}$ to be a labeled set. We use the \texttt{BIO} schema \citep{ramshaw1999text}, where the first token of an entity with type \texttt{X} is labeled as \texttt{B-X}; the consecutive tokens of the entity are labeled as \texttt{I-X}; the non-entity tokens are labeled as \texttt{O}. An NER model $\hat{\boldsymbol{y}}=f(\boldsymbol{x}; \boldsymbol{\theta})$ takes a sentence $\boldsymbol{x}$ as input and outputs a predicted label sequence $\hat{\boldsymbol{y}}$, where $\boldsymbol{\theta}$ is its parameter set. We train the NER model by minimizing the following loss,
\begin{equation}
    \mathcal{L} = \frac{1}{|D|}\sum_{i=1}^{|D|} \ell(\boldsymbol{y}_i, f(\boldsymbol{x}_i;\boldsymbol{\theta})),
\end{equation}
where $\ell(\cdot, \cdot)$ can be the cross-entropy loss for token-wise classification model or negative likelihood for CRF model \citep{lafferty2001conditional, chu2019accounting}.

Following \citet{meng2021distantly}, we build the NER model upon the RoBERTa model \citep{liu2019roberta} by adding prediction heads. Specifically, we set an entity head $f^e$ to predict whether a given token belongs to an entity and also a classification head $f^c$ to predict the class of a given token. Both heads take the contextualized representations from a RoBERTa encoder. We decompose the original label sequence $\y$ into a sequence of binary span labels $\boldsymbol{e}$ and a sequence of category labels $\boldsymbol{c}$. The span labels are obtained by transforming \texttt{B-X} and \texttt{I-X} into positive labels (denoted as \texttt{E}), and \texttt{O} labels are remained as negative labels. In $\boldsymbol{c}$, only non-empty tokens have category labels (i.e., \texttt{B-X} and \texttt{I-X}). The entity head $f^e$ is trained on $\boldsymbol{e}$ with the binary cross-entropy loss, while the classification head $f^c$ is trained on $\boldsymbol{c}$ with the cross-entropy loss. This model design allows us to handle span and category errors with distinct treatments, further details of which will be provided in Section \ref{sec:co-training}.

In inference, entities are first identified based on the outputs from the entity head, which are then classified using the classification head. The procedure is formalized as,
\begin{equation}
    \hat{y}=\left\{\begin{matrix}
 \texttt{O}, & f^e(x)\leq t \\ \,\arg\max f^c(x), & f^e(x) > t
\end{matrix}\right.\,,
\end{equation}
where $t$ is the threshold for entity identification, which is set to 0.5 by default.

\subsection{NER with Noisy Data}
In the noisy NER setting, the labels in $D$ are typically collected via crowdsourcing \citep{rodrigues2018deep, finin2010annotating} or with distant supervisions from knowledge bases \citep{liang2020bond}, which wrongly recognize many entities and often provide wrong categories for entities. Interartive self-training has proven effective in improving NER performance when learning from noisy data \citep{liang2020bond, meng2021distantly}: the predicted label sequence $\boldsymbol{\hat{y}}_i$ from the current model iteration serves as pseudo labels for the subsequent iteration,
\begin{equation}
    \mathcal{L}_{\text{Self}} = \frac{1}{|D|}\sum_{i=1}^{|D|} \ell(\boldsymbol{\hat{y}}_i, f(\boldsymbol{x}_i;\boldsymbol{\theta})).
    \label{eq:self-training}
\end{equation}
In this paper, we also require a small guidance set $C$, the labels of which are examined by domain experts to ensure high quality. Typically, we only require $|C| \ll |D|$. It is both affordable and practical to obtain a small set of high-quality data while collecting large-scale noisy data via crowdsourcing or distant supervision. In Section \ref{sec:method}, we will introduce our Self-Cleaning framework to guide the noisy NER learning with the guidance set.

\begin{figure}
\centering
\begin{subfigure}{.24\textwidth}
  \centering
  \includegraphics[height=3.15cm]{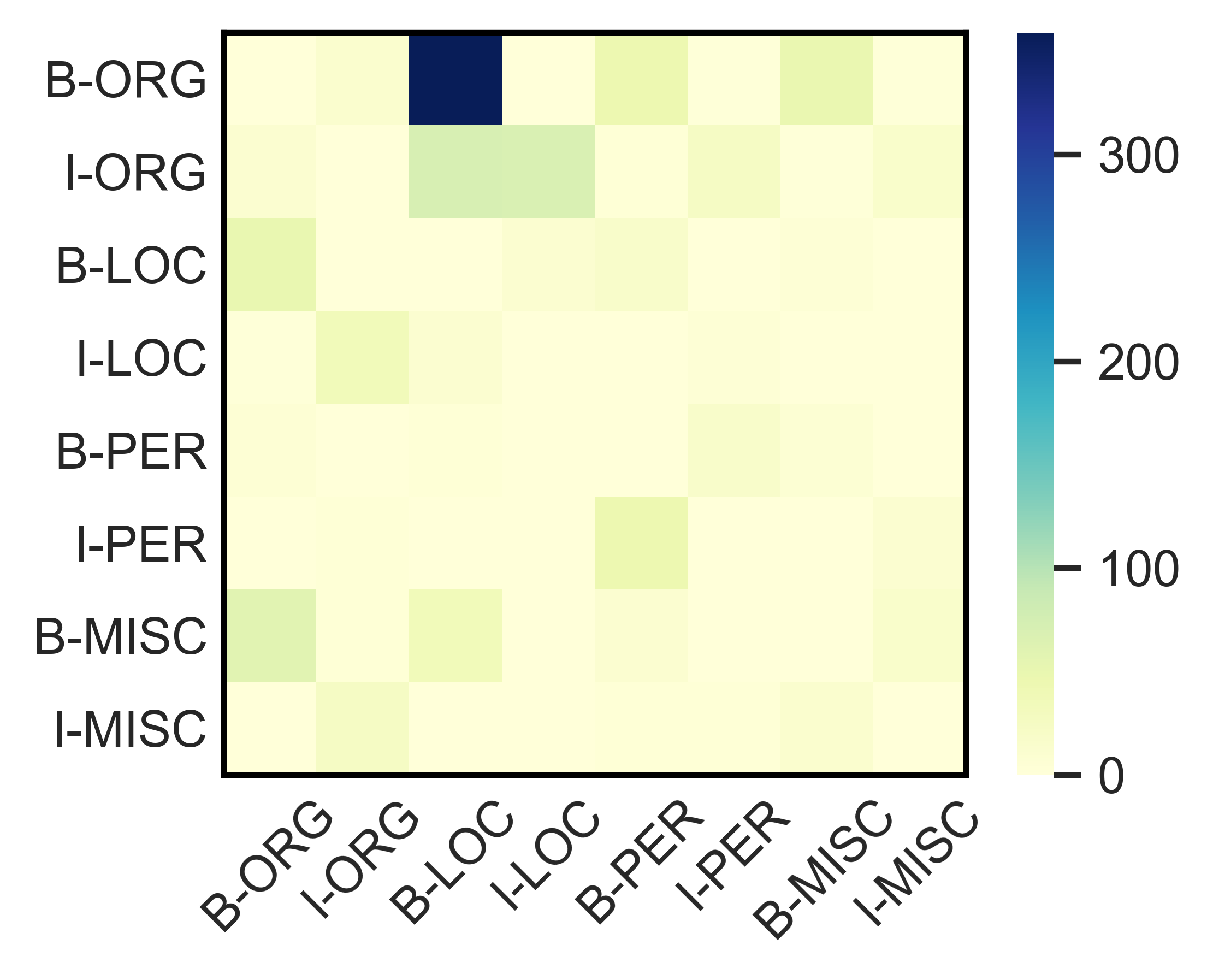}
  \caption{Crowdsourcing.}
  \label{fig:labelme}
\end{subfigure}%
\begin{subfigure}{.24\textwidth}
  \centering
  \includegraphics[height=3.15cm]{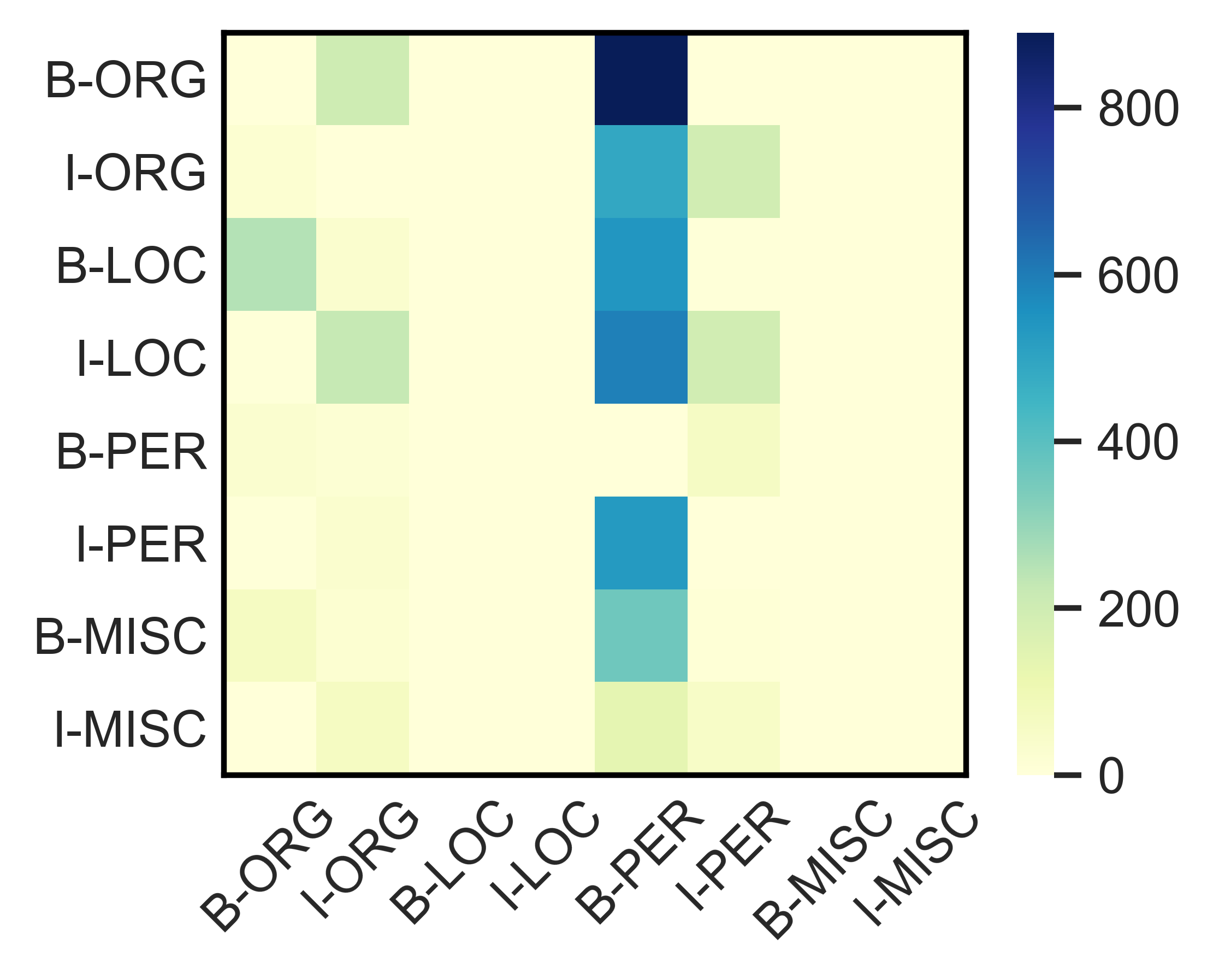}
  \caption{Distant supervision.}
  \label{fig:music}
\end{subfigure}
\caption{Confusion matrices of CoNLL03 with crowdsourcing labels and distant supervisions. The x-axis refers to noisy labels while the y-axis are ground-truth labels. The value of each entry is the frequency of this confusion pair (e.g., mistakenly label \texttt{B-LOC} as \texttt{B-ORG}).}
\vspace{-4mm}
\label{fig:noise_anlysis}
\end{figure}

\subsection{Noise Pattern Analysis}
\label{sec:analysis}
We investigate the noise patterns on the CoNLL03 dataset with crowdsouring labels collected by \citet{rodrigues2014sequence} and distant supervisions collected by \citet{liang2020bond}. We find two types of errors: (1) {\textbf{Span error}}, where the span of the entity is not correctly recognized. For example, an error would occur if only \texttt{Arizona} was recognized in \texttt{Arizona State University}. The wrong entity span could either be shorter or longer than the span of the ground-truth entity. (2) \textbf{Category error}, where the entity is assigned an incorrect category.\footnote{In the rest of the paper, we use the terms \emph{class} and \emph{category} interchangeably.} An example of this would be labeling \texttt{Arizona State University} as a \texttt{location}.

We first calculated the proportion of entity spans that overlap with but do not perfectly match the ground-truth entity: it is \textbf{11.0\%} for the crowdsourcing dataset and \textbf{12.8\%} for the distant supervision dataset, a considerable amount of error that is likely to affect the resulting model. To analyze category errors, we present the confusion matrices on two datasets in Figure \ref{fig:noise_anlysis}.\footnote{The values of diagonal entries corresponding to correct labels are set to 0, otherwise the noise patterns in the non-diagonal entries are invisible. There are several crowdsourced annotations for each token, so we aggregate them into one label using majority voting.} 
In the crowdsourcing dataset, \texttt{ORG} is often mislabeled as \texttt{LOC}, because the CoNLL dataset contains sports news in which team home cities or countries (locations) are also used as the name of the team (organizations),
which easily confuses naive annotators. And due to the entity ambiguity in knowledge bases, all the classes could be mislabeled as \texttt{PER} in the distant supervision dataset, especially \texttt{ORG}. 

Finally, a substantial proportion of ground-truth entities, \textbf{28.9\%} and \textbf{25.3\%}, are missing from the crowdsourcing and distant supervision datasets respectively. This finding underscores the importance of self-training, a crucial technique in previous noisy NER learning methods \citep{meng2021distantly, liu2021noisy,liang2020bond}, as it allows pseudo labels to recover these missing entities. However, in the absence of appropriate guidance, these pseudo labels may perpetuate both span and category errors. These errors, in turn, could be amplified due to the confirmation bias \citep{tarvainen2017mean, arazo2020pseudo, chen2019cn}, leading to a decline in performance.

\begin{figure*}
    \centering
    \includegraphics[width=15.8cm]{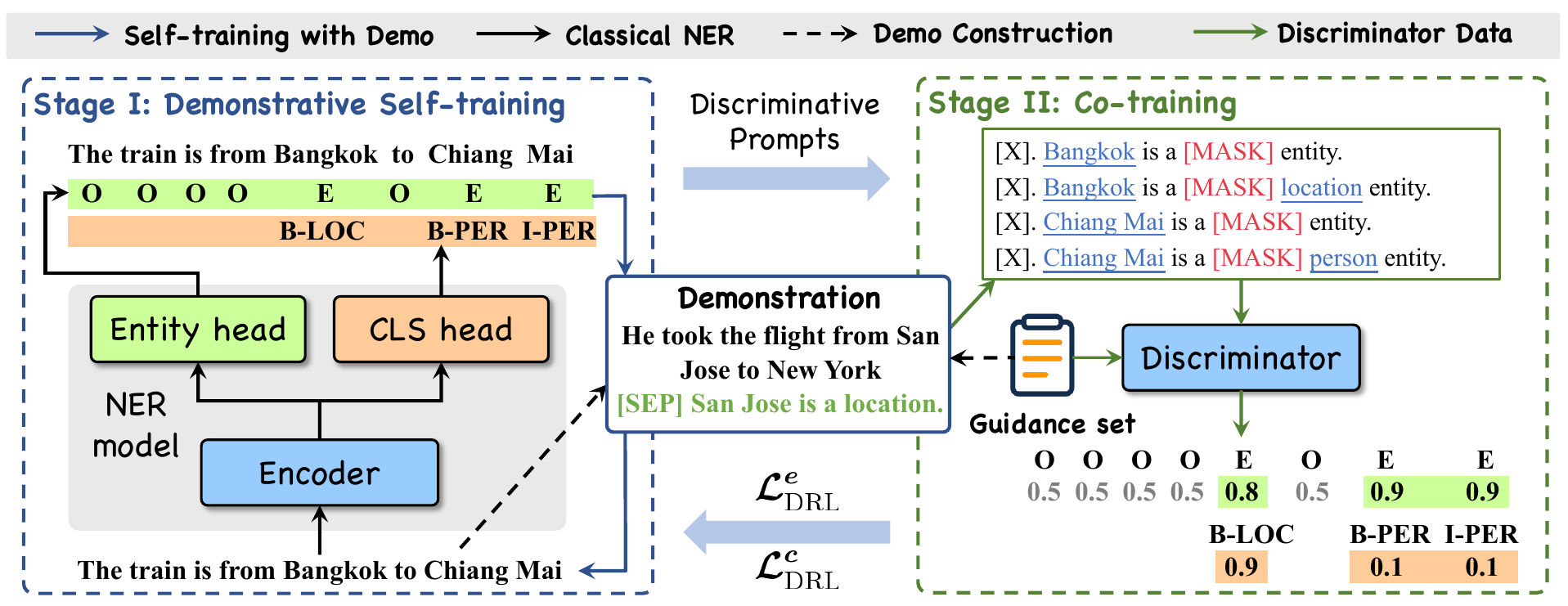}
    \caption{Overview of Self-Cleaning. In Stage I, we use the entities in the guidance set as clean demonstrations to augment the NER model's training. In Stage II, the discriminative prompts is filled with the predictions of the NER model, and then input into the prompt-based discriminator. The NER model is updated by Eq.(\ref{eq:wce}) with the weights $\boldsymbol{w}^e$ and $\boldsymbol{w}^c$ provided by the discriminator. Conversely, the high-quality pseudo labels of the NER model are used to fine-tune the discriminator.  In Stage III, we fine-tune the obtained NER model on the guidance set.\vspace{-5mm}}
    \label{fig:framework}
\end{figure*}

\section{Method: Self-Cleaning}
\label{sec:method}

In this section, we introduce Self-Cleaning in detail. We begin with the key component of Self-Cleaning: the prompt-based discriminator, explained in Section \ref{sec:discriminator}.  We then present the training procedure (as shown in Figure \ref{fig:framework}), which consists of three stages: 

\noindent \textbf{Stage I: Demonstrative self-training.} In this stage, high-confidence predictions from the NER model are used as pseudo labels to iteratively refine itself, a process often referred to as self-training \citep{liang2020bond, meng2021distantly}. To enhance the robustness of the pseudo labels, we propose a mechanism called clean demonstration, in which entities from the guidance set serve as demonstrations to elicit robust predictions from the NER model. Details of the clean demonstration mechanism can be found in Section \ref{sec:demo}.

\noindent \textbf{Stage II: Co-training.} In this stage, we introduce a co-training strategy to fine-tune the NER model and the discriminator in a collaborative manner. Specifically, the discriminator's outputs are employed to guide the NER model's training by reweighting the training labels, while high-quality predictions from the NER model are chosen to augment the guidance set used for the discriminator's training. Details of the co-training strategy and the criteria for evaluating the quality of predictions are provided in Section \ref{sec:co-training}.

\noindent \textbf{Stage III: Fine-tuning.} To further improve the performance, we fine-tune the obtained NER model only with the guidance set.

\subsection{PLM as a Unified Discriminator}
\label{sec:discriminator}
In \model{}, we use a discriminator $g$ aims to evaluate the accuracy of assigned labels to guide the NER model's training. The rationale is that labels with low accuracy should be downweighted to mitigate their influence during model training, while the accurate labels should be retained.

We identified two error types in the noisy NER data in Section \ref{sec:analysis} which can be straightforwardly modeled by the descriminaror: span error and category error. Instead of training two separate discriminators to handle each type of error, we propose to train a unified discriminator using error-type-specific prompts to elicit different outputs. This approach not only saves memory space, but also leverages the power of prompt tuning, which has been shown to effectively utilize the knowledge embedded in the parameters of pre-trained language models (PLMs) \citep{brown2020language, li2020survey}. With prompt tuning, it is possible to learn an effective discriminator with a small guidance set. In the following, we use RoBERTa \citep{liu2019roberta} as the backbone model of the discriminator and respectivly prepare Masked Language Model (MLM) style prompts. It is important to note that other PLMs, such as generative language models \citep{radford2019language}, could be seamlessly integrated into our framework by modifying the prompts accordingly. We design the following two types of discriminative prompts,
\begin{itemize}[leftmargin=*]
    \item \textbf{Span}: \texttt{[X]}. \texttt{[Y]} is a \texttt{[MASK]} entity.
    \item \textbf{Category}: \texttt{[X]}. \texttt{[Y]} is a \texttt{[MASK]} \texttt{[Z]} entity.
\end{itemize}
\texttt{[X]} is the placeholder for a sentence $\boldsymbol{x}$, \texttt{[Y]} is the placeholder for an entity $e$ and \texttt{[Z]} is the placeholder for a class $c$. The discriminator is trained to fill \texttt{correct} in the \texttt{[MASK]} token when the entity/class is appropriate given the context sentence, and \texttt{wrong} otherwise. The discriminative score of the evaluated entity or class is given by
\begin{align*}
    \begin{split}
        w^e(e) &= P_S(\texttt{correct}|\texttt{[X]}=\boldsymbol{x}, \texttt{[Y]}=e)\,, \\
        w^c(c) &= P_C(\texttt{correct}|\texttt{[X]}=\boldsymbol{x}, \texttt{[Y]}=e, \texttt{[Z]}=c)\,,
    \end{split}
\end{align*}
where $P_S$ and $P_C$ represent the probability associated with the span prompt and the category prompt, respectively. Given a sentence and its label sequence, we extract the entities in it and their corresponding classes from contiguous spans with \texttt{B-X} and \texttt{I-X} labels in the data. For example, given \texttt{[San Jose is a city]} and \texttt{[B-LOC, I-LOC, O, O, O]}, \texttt{San Jose} will be extracted as an entity and its class is \texttt{LOC}. We transform \texttt{LOC} and other category names into a meaningful word \texttt{location} which would fit naturally in a sentence. The details of the conversion can be found in Section \ref{app:verb}.

We first pre-train the discriminator to ensure a good starting point, treating the entities in the guidance set as positive samples. In this context, we will abuse the notation $C$ to denote the set of positive samples drawn from the guidance set.  We create incorrect entities and labels via data augmentations. Unlike in classification scenarios involving noisy label learning \cite{han2018co}, simulating noisy NER labels has to also provide negative examples for span errors. We investigated the datasets used in Section \ref{sec:analysis} and found that around 80\% span-error entities deviate from the ground-truth entities by a single word. Thus, we create negative entities by randomly adding or removing a word around entities in the guidance set. For example, we transform \texttt{Arizona State University} into \texttt{State University} as a negative entity. For category errors, we randomly flip the classes of entities in the guidance set. We denote the set of negative samples as $B$. The discriminator is trained to minimize the following loss function,
\begin{align*}
    \mathcal{L}_w&=-\mathbb{E}_{e, c\sim C} \big [\log w^e(e) + \log w^c(c) \big ] \\
     &-\mathbb{E}_{\Tilde{e}, \Tilde{c}\sim B}\big [\log (1 - w^e(\Tilde{e})) + \log (1-w^c(\Tilde{c}))\big ],
\end{align*}
where $1 - w^e(\Tilde{e})$ and $1 - w^c(\Tilde{c})$ are essentially $P_S(\texttt{wrong}|\cdot, \Tilde{e})$ and $P_C(\texttt{wrong}|\cdot, \Tilde{c})$. 

\subsection{Stage I: Demonstrative Self-training}
\label{sec:demo}
In this stage, we employ a self-training strategy enriched by demonstrations, to improve the performance of the NER model. Prior research \citep{zhang2022robustness, lee2021good} has established that demonstrations can boost the robustness of PLMs. Consequently, we propose to incorporate clean entities from the guidance set into the input of the NER model to stimulate more robust outputs. These enhanced outputs are then used as pseudo labels for self-training, as specified in Eq.\eqref{eq:self-training}.

Technically, we follow the \texttt{instance-oriented} method in \citep{lee2021good} to find demonstrations. For each sentence in the noisy training set, we first retrieve similar sentences from the guidance set using \texttt{SBERT} scores \citep{reimers2019sentence}. Then, the entities in the retrieved guidance sentences are used to form the clean demonstrations $\boldsymbol{\Tilde{x}}$, which are appended as additional tokens to the original training sentence $\boldsymbol{x}$. The inputs of the NER model become $[\boldsymbol{x} ; \boldsymbol{\Tilde{x}}]$. For example in Figure \ref{fig:framework}, \texttt{San Jose-LOC} is used to form the clean demonstration $\boldsymbol{\Tilde{x}}=$\texttt{[SEP] San Jose is a location}. During inference, we empirically found that demonstrations did not improve performance, hence we only input the original sentence $\x$ into the NER model.

\subsection{Stage II: Co-training}
\label{sec:co-training}
In this stage, we fine-tune the NER model $f$ and the discriminator $g$ in a collaborative manner to improve the performance of both. On the one hand, the discriminator guides the NER model's training by reweighting the training labels. On the other hand, the high-quality pseudo labels generated by the NER model are used to augment the discriminator's training.

\noindent\textbf{Discriminator-guided training for NER.} Even though pseudo labels can effectively improve the performance,  they may reproduce the noise present in the noisy training set, leading to confirmation bias \cite{tarvainen2017mean} that impedes further model improvement. Therefore, we propose using the discriminator to guide self-training by reweighting the pseudo labels. As shown in Figure \ref{fig:framework}, we first extract the pseudo entities and their corresponding classes from the pseudo label sequences, and then insert them into the discriminative prompts. The outputs of the discriminator are used as weights for the pseudo labels, resulting in the following discriminative reweight loss (DRL), 
\begin{equation}
    \label{eq:wce}
    \mathcal{L}^{e/c}_{\text{DRL}} = -\frac{1}{|D|}\sum_{i=1}^{|D|}\sum_{j=1}^n w_{ij}^{e/c}\log f^{e/c}_{\hat{e}_{ij}/\hat{c}_{ij}}(x_{ij};\boldsymbol{\theta}),
\end{equation}
where $\hat{e}_{ij}$ and $\hat{c}_{ij}$ denote the pseudo span labels and category labels, respectively, for the $j$-th token in the $i$-th sentence; and $f^{e/c}_{\hat{e}_{ij}/\hat{c}_{ij}}$ refers to the entry of $\hat{e}_{ij}$ or $\hat{c}_{ij}$ in the corresponding probability distribution. Note that an entity could consist of several tokens, to which we allocate equivalent weights. We set the weights of negative span labels \texttt{O} to 0.5 by default to avoid overfitting on them.

\noindent\textbf{Enhancing discriminator with high-quality pseudo labels.} Conversely, we use high-quality pseudo labels generated by the NER model to enhance the discriminator's training. We assess the quality of pseudo labels based on two criteria: \emph{accuracy} and \emph{informativeness}. Firstly, We follow \citet{yao2021jo} to use Jensen-Shannon divergence (JSD) as a proxy to evaluate the accuracy of the pseudo labels of a token $x_i$,
\begin{align*}
    \begin{split}
            q(\hat{e}_i) &= 1 - \text{JSD}\big (f^e(x_i) \parallel \texttt{one\_hot}(e_i)\big ), \\
    q(\hat{c}_i) &= 1 - \text{JSD}\big (f^c(x_i) \parallel \texttt{one\_hot}(c_i)\big ),
    \end{split}
\end{align*}
where $\hat{e}_i$ and $\hat{c}_i$ are pseudo span label and category label for token $x_i$, while $f^e(x_i)$ and $f^c(x_i)$ are their corresponding probabilities from the entity head and classification head. $e_i$ and $c_i$ are observed labels in the training set, which are transformed into distributions by one-hot encoding.\footnote{Label smoothing is used to avoid 0 entries.} However, the \emph{mostly correct} pseudo labels selected by the above metric is not always helpful for the discriminator training, as they may not carry new information. Intuitively, if the discriminator shows uncertainty for its own prediction, that particular pseudo label becomes more informative. Similar to active learning \cite{chu2021improve, schroder2021revisiting}, we identify the most informative samples using the prediction entropy of the discriminator as a measure of uncertainty. The resulting token-level selection score $s(\cdot)$ is defined as,
\begin{align*}
    \begin{split}
        s(\hat{e}_i) = \mathbb{H}\big (w^e(\hat{e}_i)\big ) \cdot q(\hat{e}_i)\,, \\
        s(\hat{c}_i) = \mathbb{H}\big (w^c(\hat{c}_i)\big ) \cdot q(\hat{c}_i)\,,
    \end{split}
\end{align*}
where $\mathbb{H}$ is the entropy function while $w^e(\hat{e}_i)$ and $w^c(\hat{c}_i)$ are discriminative scores of pseudo labels. However, our discriminator works at the entity, not token, level. We form the entity-level selection score by averaging the token-level scores within an entity, $\frac{1}{L}\sum_{i}^{L}s(\hat{e}_i)$ and $\frac{1}{L}\sum_{i}^{L}s(\hat{c}_i)$, where $L$ is the number of tokens in the entity. We select top-$K$ entities as pseudo positive samples for discriminator, where $K$ is set as a hyper-parameter. For each pseudo positive samples, we simulate pseudo negative samples in the same way as described in Section \ref{sec:discriminator} to facilitate the discriminator training. To improve the few-shot ability of the discriminator \citep{gao2021making}, we use the approach described in Section \ref{sec:demo} to generate demonstrations for the discriminator's inputs when fine-tuning and utilizing the discriminator in the co-training stage.

Lastly, in Stage III, we further fine-tune the obtained NER model only with the guidance set, as suggested in \citet{jiang2021named}.

\section{Experiments}
\begin{table}[htp!]
\centering
\caption{Dataset statistics.}
\begin{tabular}{@{}lccc@{}}
\toprule
\textbf{Dataset} & \textbf{\#Types} & \textbf{\#Train} & \textbf{\#Test} \\ \midrule
\textbf{CoNLL03-C}        & 4                & 5,985            & 3,453           \\ \midrule
\textbf{CoNLL03}          & 4                & 14,041           & 3,453           \\
\textbf{OntoNotes}     & 18               & 59,924           & 8,262           \\
\textbf{Wikigold}         & 4                & 1,142            & 274             \\ \bottomrule
\end{tabular}
\end{table}
\subsection{Experiment Setup} 
\textbf{Datasets.} We conduct the experiments on two kinds of noisy English NER datasets:

\noindent\textbf{Crowdsourcing.} We use a crowdsourced NER dataset \citep{rodrigues2018deep} based on CoNLL03, denoted as CoNLL03-C, where 5,985 sentences are labeled by 47 non-expert annotators.  Redundant crowdsourced annotations for each token are aggregated into a single noisy label using majority voting.

\noindent\textbf{Distant supervision.} We use three benchmarks for distant supervision datasets including CoNLL03 \citep{sang2003introduction}, OntoNotes5.0 \citep{weischedel2013ontonotes} and Wikigold \citep{balasuriya2009named}. We follow BOND \citep{liang2020bond} to obtain distant supervisions using existing knowledge bases. The noise in these datasets is more systematic, as it is mainly caused by entity ambiguity or missing entities.

We randomly sample the small guidance set from the training set with ground-truth labels, ensuring that all types are covered in the guidance set at least $\left \lfloor \frac{|C|}{\# Types} \right \rfloor$ times. We use guidance sets of 200, 500, and 50 sentences on CoNLL03 and CoNLL03-C, OntoNotes5.0, and Wikigold, respectively. For each dataset the guidance sets are less than 5\% of the size of the full set. The size of the guidance set $C$ is an important hyperparameter that impacts the final performance, so we further study its influence in Section \ref{sec:inf-guidance}. We use \texttt{roberta-base} as the backbone model for both the NER model and the discriminator. More implementation details can be found  in Appendix \ref{app:implementation}. We also conduct a comprehensive study of different model designs in Appendix \ref{app:model_design}, including using generative language models \citep{chung2022scaling} as discriminator backbones and different combinations of backbone models for the NER model and the discriminator.

\begin{table}[htp!]
\centering
\caption{Results on CoNLL03-C.}
\label{tab:crowd}
\resizebox{.43\textwidth}{!}{
\begin{tabular}{@{}lccc@{}}
\toprule
\multicolumn{1}{c}{\textbf{Methods}} & \textbf{Pre.} & \textbf{Rec.} & \textbf{F1} \\ \midrule
\textbf{Distant RoBERTa}             &       0.824   &      0.796    &      0.805       \\
\textbf{BOND}                        &       0.775   &      0.806    &      0.787   \\
\textbf{RoSTER}                      &       0.790   &      0.822    &      0.804  \\ \midrule
\textbf{Fine-tune RoBERTa} & 0.695 &  0.699 & 0.694  \\
\textbf{Fine-tune RoSTER}            &       0.778   &      0.831    &      0.802  \\
\textbf{NEEDLE}                      &       0.822   &      0.863    &    0.842 \\
\textbf{GLC}                         &       0.803   &      0.791    &    0.790  \\
\textbf{Meta-Reweight}               &       0.768   &      0.835    &    0.799  \\
\textbf{Self-Cleaning}                   &        \textbf{0.849}  &      \textbf{0.876}    &      \textbf{0.862}    \\ \bottomrule
\end{tabular}
}
\vspace{-4mm}
\end{table}

\begin{table*}[!htp]
\vspace{-4mm}
\caption{Results on distant supervision NER datasets. $p$-value is reported to show the statistical significance.}
\label{tab:ds}
\centering
\resizebox{.95\textwidth}{!}{
\begin{tabular}{@{}lccccccccc@{}}
\toprule
\multicolumn{1}{c}{\multirow{2}{*}{\textbf{Methods}}} & \multicolumn{3}{c}{CoNLL03}                 & \multicolumn{3}{c}{OntoNotes5.0}            & \multicolumn{3}{c}{Wikigold}                \\
\multicolumn{1}{c}{}                                  & \textbf{Pre.} & \textbf{Rec.} & \textbf{F1} & \textbf{Pre.} & \textbf{Rec.} & \textbf{F1} & \textbf{Pre.} & \textbf{Rec.} & \textbf{F1} \\ \midrule
\textbf{Distant RoBERTa}                             & 0.784         & 0.756         & 0.743       & 0.760         & 0.715         & 0.737       & 0.534         & 0.623         & 0.566       \\
\textbf{BOND}                                         & 0.849   &  0.854   &  0.848       & 0.740         & 0.767         & 0.753       & 0.541  &   0.679  &   0.595      \\
\textbf{RoSTER}                                       & 0.856         & 0.867         & 0.859       & 0.759         & 0.792         & 0.771       & 0.581         & 0.716         & 0.637       \\ \midrule
\textbf{Fine-tune RoBERTa} & 0.695 & 0.699 &  0.694 & 0.744 & 0.822 &    0.779 & 0.493 &    0.551   &  0.509 \\
\textbf{Fine-tune RoSTER}  & 0.850     &       0.872        &  0.860     &      0.756       &        0.797       &     0.773
          &     0.620  &  0.755  &      0.675       \\
\textbf{NEEDLE}  & 0.861   & 0.877 & 0.866  & 0.730 & 0.782 & 0.751 & 0.707  & 0.777  & 0.738   \\
\textbf{GLC}  & 0.866  &   0.853  & 0.856  & - & - & - & 0.626  & 0.754 & 0.679 \\
\textbf{Meta-Reweight} & 0.839 & 0.866 & 0.851  & 0.737 & 0.781 & 0.755 & 0.609 & 0.746 & 0.665 \\
\textbf{Self-Cleaning}  &      \textbf{0.883}     &     \textbf{0.882}   &       \textbf{0.882}   &   \textbf{0.809} &  \textbf{0.846} &  \textbf{0.826}  &       \textbf{0.761 } &   \textbf{0.798}   &  \textbf{ 0.778}    \\ \midrule
\textbf{RoBERTa (Gold)} &  0.907 & 0.930 & 0.918 & 0.884 & 0.912 & 0.897 & 0.823 & 0.858 & 0.839  \\
$p$-\textbf{value} & - & - & $<$0.005 & - & - & $<$0.001 & - & - & $<$0.001  \\
\bottomrule
\end{tabular}
}
\vspace{-4mm}
\end{table*}

\noindent\textbf{Baselines.} We compare against two broad classes of related solutions as baselines. The first class of baselines is approaches that only use noisy labels and no clean data whatsoever:
\begin{itemize}[leftmargin=*]
    \item \textbf{Distant RoBERTa}, where a RoBERTa model is fine-tuned using noisy labels.
    \item \textbf{BOND} \citep{liang2020bond} fine-tunes a RoBERTa model on noisy labels with early-stopping, and then self-trains the resulting model.
    \item \textbf{RoSTER} \citep{meng2021distantly} combines a noise-robust loss and ensemble training to improve robustness on noisy NER data, and then utilizes a language model augmented self-training.
\end{itemize}

The second class of baselines covers approaches that similar to Self-Cleaning also utilize a guidance set when training, but the guidance set is used in different ways: 

\begin{itemize}[leftmargin=*]
    \item \textbf{Fine-tune RoBERTa}, where a RoBERTa model is fine-tuned on the guidance set.
    \item \textbf{Fine-tune RoSTER}, where the final model of RoSTER is fine-tuned on the guidance set.
    \item \textbf{NEEDLE} \citep{jiang2021named} estimates the confidence scores of pseudo labels in the self-training stage using the histogram binning heuristic.
    \item \textbf{GLC} \citep{hendrycks2018using} estimates a class-level confusion matrix using the guidance set, which is used to calibrate the loss on noisy labels.
    \item \textbf{Meta-Reweight} \citep{wu2022self, shu2019meta} uses a bi-level optimization framework to learn label weights. It learns the weights of pseudo labels by minimizing the meta-loss on the guidance set in the upper level and updates the NER model with the weights in the lower level.
\end{itemize}

\subsection{Main results}

 We report the results on CoNLL03-C in Table \ref{tab:crowd} and three distant supervision datasets in Table \ref{tab:ds}, where Self-Cleaning outperforms all baselines significantly. The performance of the second group of methods is generally better than the first group, which shows the necessity of the guidance from clean data. GLC and Meta-Reweight are directly borrowed from the Machine Learning community;\footnote{GLC on OntoNotes5.0 is not reported due to its poor performance.} both of them fail to improve the performance with the guidance set. GLC estimates a confusion matrix of labels using the guidance set. However, in the NER scenario, label-level confusion is not meaningful, e.g., all non-empty labels can be labeled as \texttt{O} due to span error. NEEDLE uses the guidance set to estimate the confidence scores by heuristics. In contrast, informed by the analysis of noise that we presented, we design a discriminator to handle two types of errors in Self-Cleaning, which has shown to be a more effective way to provide guidance in the learning on noisy NER data. Please refer to Appendix \ref{app:case_study} for a detailed case study elucidating the workings of both the NER model and the discriminator in \model{}.

\begin{figure}[htp!]
    \centering
    \includegraphics[width=5cm]{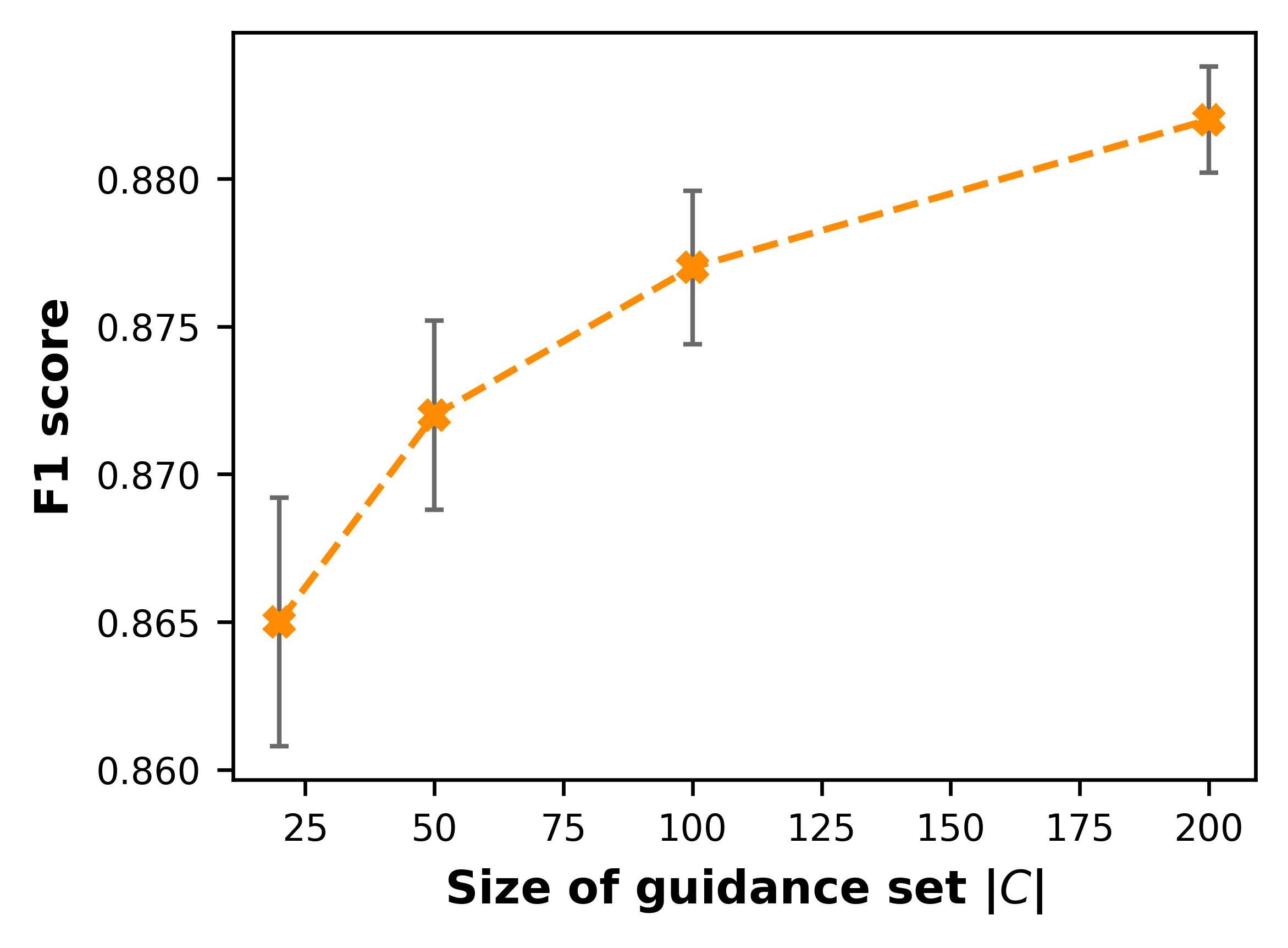}
    \caption{Results with different $|C|$.}
    \label{fig:guidance_size}
    \vspace{-3mm}
\end{figure}

\subsection{Influence of $|C|$}
\label{sec:inf-guidance}
The size of the guidance set influences the quality of the discriminator, and thus affects the final performance. We study the performance of Self-Cleaning with different sizes of guidance sets. For each $|C|$, we randomly sample 3 guidance sets. Due to the space limit, we report the mean and standard deviation of F1 score on CoNLL03, similar observations were also obtained on other datasets.

We show the results in Figure \ref{fig:guidance_size}. With a smaller guidance set, the performance of Self-Cleaning drops as the quality of discriminator gets worse. Also, the performance becomes more unstable with smaller guidance sets, since the pattern distribution in different sets is different. The results show that the selection of the guidance set is crucial to the final performance. If the guidance set is of low quality or too small, the quality of the discriminator will be the bottleneck of the final performance. 

In additional experiments  we find that to reach the compareable performance (an F1 score of 0.880) as Self-Cleaning ($|C|=200$), RoBERTa (Gold) needs 1,000 clean instances, five times more than Self-Cleaning. Directly fine-tuning a RoBERTa  model only on the same guidance set $C$ results in markedly worse performance as seen in Table \ref{tab:ds}. Noisy labels do effectively improve the sample efficiency of clean data.

\subsection{Ablation Study}
To evaluate the individual contributions of different components in Self-Cleaning, we conduct an ablation study and create the following variants: Firstly, we remove the span prompts and only reweight the category labels. Secondly, we remove the category prompts, which means the discriminator can only reweight the binary span labels.  Thirdly, we remove clean demonstrations. Lastly, we remove the co-training strategy, and only use the pre-trained discriminator. Additionally, we also report the results on Stage I and Stage II.

We present the results in Table \ref{tab:ablation}. As we discussed in Section \ref{sec:analysis}, span error is a severe issue in the noisy NER data. Without the ability to detect span errors, the performance drops considerably. Also, without clean demonstrations, the few-shot ability of the discriminator is limited and the NER model lacks of guidance when generating pseudo labels, causing a drop in performance. By comparing the results of Stage I and RoSTER, we also observe that utilizing clean demonstrations leads to an improvement in the robustness of self-training. The co-training strategy is important to improve the discriminator, covering more patterns by involving pseudo positive labels from the NER model. Lastly, the improvement from Stage II to Self-Cleaning shows that fine-tuning on the guidance set can further improve the performance.

\begin{table}[htp!]
\centering
\caption{Results of ablation study on CoNLL03.}
\label{tab:ablation}
\resizebox{.45\textwidth}{!}{
\begin{tabular}{@{}lccc@{}}
\toprule
\textbf{Methods}    & \textbf{Pre.}        & \textbf{Rec.}        & \textbf{F1}          \\ 
\midrule
w/o Span Disc.     &  0.866   &   0.885 &   0.874    \\
w/o Cat. Disc.  &   0.878 & 0.879  & 0.877                \\
w/o Demonstration   &    0.888 &   0.873  & 0.878                    \\ 
w/o Co-training  & 0.882  &   0.877  &   0.878        \\ \midrule
Stage I   & 0.861   & 0.888 &  0.874                     \\
 Stage II     & 0.881   &   0.879  & 0.880                                            \\ \midrule
\textbf{Self-Cleaning}  & 0.883  &   0.882  &   0.882
\\ \bottomrule
\end{tabular}
}
\vspace{-4mm}
\end{table}

\section{Conclusion}
In this paper, we study how to improve NER models trained on noisy labeled data with a guidance set consisting of a small number of clean instances. Our research is grounded on the noise pattern analysis on the real-world noisy NER data. We identify two NER-specific error types: span error and category error. To address these errors, we propose to use a dedicated discriminator to guide the training of the NER model. This discriminator is tailored to detect the aforementioned errors using pre-defined discriminative prompts, and its outputs are used to reweight the samples for training the NER model. We design a three-stage training procedure to unleash the power of clean instances in guiding noisy NER learning. We evaluate the proposed method on a rich set of NER benchmarks with crowdsourcing labels and distant supervisions. The results show that with a few clean instances, the proposed method can boost the performance significantly.

\section*{Limitations}
The discriminator is the key part of \model{}, however, it has several limitations. Firstly, the current version is able to handle noise within recognized entities but it falls short when dealing with noise in non-entity labels, i.e., \texttt{O} labels. Secondly, since the discriminator works at the entity level, an entity with even partially incorrect labels is fully downweighted. This approach could lead to the discarding of potentially useful labels, especially when category labels are very sparse. Future work might consider the development of discriminators that operate on the token level.

Additionally, it is also worth noting that in the current discriminator design, we did not make explicit assumptions about the underlying mechanisms generating span and categorical errors. The negative samples are simulated by randomly modifying tokens within entities and flipping their classes. Such negative samples may not be informative enough to capture the salient patterns needed to distinguish correct labels from incorrect ones, thereby limiting the final performance. For a more comprehensive understanding and identification of the root causes of errors in noisy NER data, future work might incorporate more advanced error modeling techniques, such as lexical analysis or cross-validation against external knowledge bases.

\section*{Ethics Statement}
Learning from noisy NER data diminishes the necessity for large-scale, high-quality labeled data, thereby facilitating its use in domains where obtaining expert knowledge is costly, such as in legal and financial sectors. Our proposed method paves the way for achieving a model with reasonable performance while keeping the cost of expert-labeled data within an acceptable range. It has the potential to lower the entry barrier for novices who have limited data at their disposal.

However, we should notice our method makes it easier to attack the modeling training by poisoning the guidance set. Given the limited size of the guidance set, a subtle change could drastically alter its distribution, potentially leading to the collapse of the entire training pipeline.

\bibliography{anthology,custom}
\bibliographystyle{acl_natbib}

\appendix
\newpage
\section{Implementation Details} 
\label{app:implementation}
In Self-Cleaning, we use \texttt{roberta-base} as the backbone for both NER model and the discriminator. We use AdamW optimizer to optimize both NER model and the discriminator. We pre-train the discriminator with a learning rate $2e^{-5}$ and $5e^{-6}$ during co-training. The training batch size is fixed as 64. To update the NER model, we use learning rate $5e^{-6}$ for CoNLL03 and Wikigold, and $5e^{-7}$ for OntoNotes5.0. During co-training, we choose $K=20$ pseudo entities per class to fine-tune the discriminator. We first use the noise-robust loss and ensemble training in \citet{meng2021distantly} to pre-train the NER model on noisy NER data, and then apply the proposed Self-Cleaning approach on the obtained model. GLC, Meta-Reweight and NEEDLE start with the same pre-trained model as Self-Cleaning. All experiments are repeated with 3 random seeds and 3 randomly sampled guidance sets. The averaged metrics are reported. We run our experiments on 2 NVIDIA GeForce RTX 2080Ti GPUs with 12 GB memory. 

\begin{table}[h]
\centering
\caption{Results on synthetic noisy CoNLL03.}
\label{tab:syn}
\begin{tabular}{@{}lcccc@{}}
\toprule
\multicolumn{1}{c}{\multirow{2}{*}{\textbf{Methods}}} & \multirow{2}{*}{\textbf{Type}} & \multicolumn{3}{c}{Noise Rate}             \\
\multicolumn{1}{c}{}                                  &                                & \textbf{0.2} & \textbf{0.4} & \textbf{0.6} \\ \midrule
\multirow{2}{*}{\textbf{RoSTER}}                      & \textbf{Span}              &     0.852         & 0.823        & 0.462        \\
                                                      & \textbf{Cat.}                  &     0.886         & 0.873        & 0.667        \\ \midrule
\multirow{2}{*}{\textbf{Self-Cleaning}}                  & \textbf{Span}                  &    0.901          & 0.897        & 0.896        \\
                                                      & \textbf{Cat.}                  &         0.899     & 0.895        & 0.864        \\ \bottomrule
\end{tabular}
\end{table}

\section{Verbalizer}
\label{app:verb}
We list the mapping between NER labels and words used in our prompt-based discriminator. 
\begin{itemize}[noitemsep,leftmargin=*,topsep=0pt]
    \item \textbf{CoNLL03}: \texttt{PER} - person, \texttt{LOC} - location, \texttt{ORG} - organization, \texttt{MISC} - other.
    \item \textbf{OntoNotes5.0}: \texttt{WORK\_OF\_ART} - work of art, 
\texttt{PRODUCT} - product,
\texttt{NORP} - affiliation,
\texttt{ORG} - organization,
\texttt{FAC} - facility, 
\texttt{GPE} - geo-political,
\texttt{LOC} - location,
\texttt{PERSON} - person,
\texttt{EVENT} - event,
\texttt{LAW} - law, 
\texttt{LANGUAGE} - language,
\texttt{PERCENT} - percent,
\texttt{ORDINAL} - ordinal,
\texttt{QUANTITY} - quantity,
\texttt{CARDINAL} - cardinal,
\texttt{TIME} - time,
\texttt{DATE} - date,
\texttt{MONEY} - money.
    \item \textbf{Wikigold}: \texttt{PER} - person, \texttt{LOC} - location, \texttt{ORG} - organization, \texttt{MISC} - other.
\end{itemize}

\section{Experiments on Synthetic Datasets}
\textbf{Settings.} We also evaluate Self-Cleaning on synthetic data, where we manually create noisy NER data. We create two kinds of datasets based on CoNLL03 with span and category errors, respectively. For each error type, we control the noise rates. For the span error, the noise rate controls the probability to add or remove a token around a ground-truth entity. For the category error, the noise rate defines the probability of the class of a ground-truth entity to be flipped into a noisy class. 

\noindent\textbf{Results.} We present the results in Table \ref{tab:syn}, where we also show the results of RoSTER to study the effect of noise rate. We can observe that with larger noise rate, the performance of RoSTER decreases significantly. But with our dedicated discriminator, both types of errors can be detected and downweighted, leading to a robust performance.

\begin{table}[htp]
\centering
\caption{Results of various discriminator backbones.}
\label{tab:discriminator}
\resizebox{.45\textwidth}{!}{
\begin{tabular}{@{}lccc@{}}
\toprule
\textbf{Backbone}    & \textbf{Pre.}        & \textbf{Rec.}        & \textbf{F1}          \\ 
\midrule
\texttt{roberta-base}  &  0.883 & 0.882 & 0.882 \\
\texttt{roberta-large} & 0.885 & 0.883 & 0.884 \\ \midrule
\texttt{flan-t5-small} & 0.878  &   0.873  & 0.875  \\
\texttt{flan-t5-base} & 0.884 & 0.877  &   0.878   \\
\texttt{flan-t5-large} & 0.889 & 0.877 & 0.881 \\ \bottomrule
\end{tabular}
}
\vspace{-2mm}
\end{table}

\section{Experiments of different model designs}
\label{app:model_design}
\subsection{Study of Discriminator Backbones}
\label{sec:backbone}
To study the effect of different kinds of discriminators, we also incorporate Self-Cleaning with Generative Language Model (GLM) based discriminator. Specifically, we use Flan-T5, an instruction fine-tuned GLM family \citep{chung2022scaling}. Accordingly, we design the following two prompts,
\begin{itemize}[leftmargin=*]
    \item \textbf{Span}: \texttt{[X]}. \texttt{[Y]} is an entity. Is it correct?
    \item \textbf{Category}: \texttt{[X]}. \texttt{[Y]} is a \texttt{[Z]} entity. Is it correct? 
\end{itemize}
The GLM-based discriminator is supposed to choose an answer from [\texttt{correct}, \texttt{wrong}]. We use the same method in Section \ref{sec:discriminator} to create create both positive and negative samples for the pre-training of the discriminator. We consider three Flan-T5 variants with varying parameter sizes to understand the impact of model scaling. Additionally, we include results obtained by using \texttt{roberta-large} as the backbone of the discriminator.

In Table \ref{tab:discriminator}, we report the results on CoNLL03. We can observe that with a larger backbone model, the final performance is slightly better. Interestingly, both MLM-based and GLM-based discriminators achieve similar final performance. The success of GLM-based discriminators make it possible to introduce more powerful GLMs like the GPT family \citep{radford2019language} in the future. However, the performance gains from larger models are marginal, suggesting a performance bottleneck. We hypothesize that the randomly generated negative samples may not be sufficiently informative. We leave how to create useful negative samples for the discriminator as an important future work.

\subsection{Study of Encoder Configurations}
In \model{}, we employ two \texttt{roberta-base} models as encoders for the NER model and the discriminator respectively. Additionally, we experimented with alternative designs, such as building both the NER model and the discriminator on top of a single \texttt{roberta-base} encoder. In this configuration, we added an entity head and a classification head for the NER model, while also incorporating an MLM head for the discriminator. We also conducted similar tests using the \texttt{roberta-large} model and have reported these results as well.

The results on CoNLL03 are reported in Table \ref{app:config}. The variants utilizing \texttt{roberta-large} show better performance than those based on \texttt{roberta-base}, owing to the increased power of the backbone model. However, when the NER model and the discriminator share a single encoder, it negatively affects the final performance. Specifically, the RoBERTa encoder, when trained on noisy NER data, tends to propagate its noise to the discriminator, thereby affecting its quality. Therefore, to ensure the isolation between clean and noisy data, we recommend employing separate encoders for the NER model and the discriminator. This design is also more flexible as we are able to use different backbone models for the NER model and the discriminator, as we did in Section \ref{sec:backbone}.

\begin{table}[htp]
\centering
\caption{Results of different encoder configurations.}
\label{app:config}
\resizebox{.45\textwidth}{!}{
\begin{tabular}{@{}lccc@{}}
\toprule
\textbf{Encoder}    & \textbf{Pre.}        & \textbf{Rec.}        & \textbf{F1}          \\ 
\midrule
one \texttt{roberta-base}  &  0.881 & 0.875 & 0.878 \\
two \texttt{roberta-base} & 0.883 & 0.882 & 0.882 \\ \midrule
one \texttt{roberta-large} & 0.889  &   0.879  & 0.884  \\
two \texttt{roberta-large} & 0.897 & 0.887  &   0.892 \\  \bottomrule
\end{tabular}
}
\end{table}

\section{Case Study and Analysis}
\label{app:case_study}
\subsection{Case Study of the NER model}
In Table \ref{tab:case}, we perform case study to understand the advantage of \textbf{Self-Cleaning} with a concrete example, by comparing with the best baseline without guidance \textbf{RoSTER} and with guidance \textbf{NEEDLE}. Without the guidance about the span and category errors, RoSTER fails to detect the correct span of \texttt{Sheffield Shield} and classify \texttt{Bellerive Oval} even though the span is correct. NEEDLE estimates the confidence scores according to the NER model's outputs via the histogram binning heuristic \cite{zadrozny2001obtaining}, which is ineffective to handle both span and category errors.  Self-Cleaning is able to downweight the noisy entities with wrong spans and classes, leading to the correct recognition of the testing sentence. 

\begin{table*}[htp!]
\centering
\caption{Case study of Self-Cleaning and baselines. The sentence is from CoNLL03.}
\label{tab:case}
\resizebox{.85\textwidth}{!}{
\begin{tabular}{@{}ll@{}}
\toprule
\multirow{2}{*}{\textbf{Ground truth}} 
                                       & Score on the first day of the four-day \textcolor{blue}{[Sheffield Shield]$_{\text{MISC}}$} match between 
                                       \\ &
                                       \textcolor{red}{[Tasmania]$_{\text{LOC}}$}  
                                       and \textcolor{red}{[Victoria]$_{\text{LOC}}$} at \textcolor{red}{[Bellerive Oval]$_{\text{LOC}}$} on Friday. 
                                       \\ \midrule
\multirow{2}{*}{\textbf{RoSTER}}       & Score on the first day of the four-day \textcolor{blue}{[Sheffield]$_{\text{MISC}}$} Shield match  between 
                                       \\ &
                                       \textcolor{red}{[Tasmania]$_{\text{LOC}}$}  
                                       and \textcolor{red}{[Victoria]$_{\text{LOC}}$} at \textcolor{orange}{[Bellerive Oval]$_{\text{ORG}}$} on Friday. \\ \midrule
\multirow{2}{*}{\textbf{NEEDLE}}       & Score on the first day of the four-day \textcolor{blue}{[Sheffield]$_{\text{MISC}}$} Shield match  between 
                                       \\ &
                                       \textcolor{red}{[Tasmania]$_{\text{LOC}}$}  
                                       and \textcolor{red}{[Victoria]$_{\text{LOC}}$} at \textcolor{orange}{[Bellerive]$_{\text{ORG}}$} Oval on Friday. \\ \midrule
\multirow{2}{*}{\textbf{Self-Cleaning}}   & Score on the first day of the four-day \textcolor{blue}{[Sheffield Shield]$_{\text{MISC}}$} match between 
                                       \\ &
                                       \textcolor{red}{[Tasmania]$_{\text{LOC}}$}  
                                       and \textcolor{red}{[Victoria]$_{\text{LOC}}$} at \textcolor{red}{[Bellerive Oval]$_{\text{LOC}}$} on Friday.\\ \bottomrule
\end{tabular}
}
\end{table*}

\subsection{Case Study of the Discriminator}
\label{app:dis}
We present some example outputs of the discriminator in Table \ref{tab:case_dis}. Even when the class of an entity is incorrectly identified, the discriminator can still evaluate the span correctly. For instance, in the first example, \texttt{China} is correctly recognized as an entity, but is misclassified as \texttt{ORG}. The discriminator accurately assigns a low score to the category label and a high score to the span label. However, if the span label is incorrect, the category label will also be downweighted by the discriminator. For example, in the third case, both the span score and category score are low. Intuitively, correct entity recognition is a prerequisite for correct classification, making it meaningless to preserve the category label if the span label is incorrect.

\begin{figure}[htp]
    \centering
    \includegraphics[width=5cm]{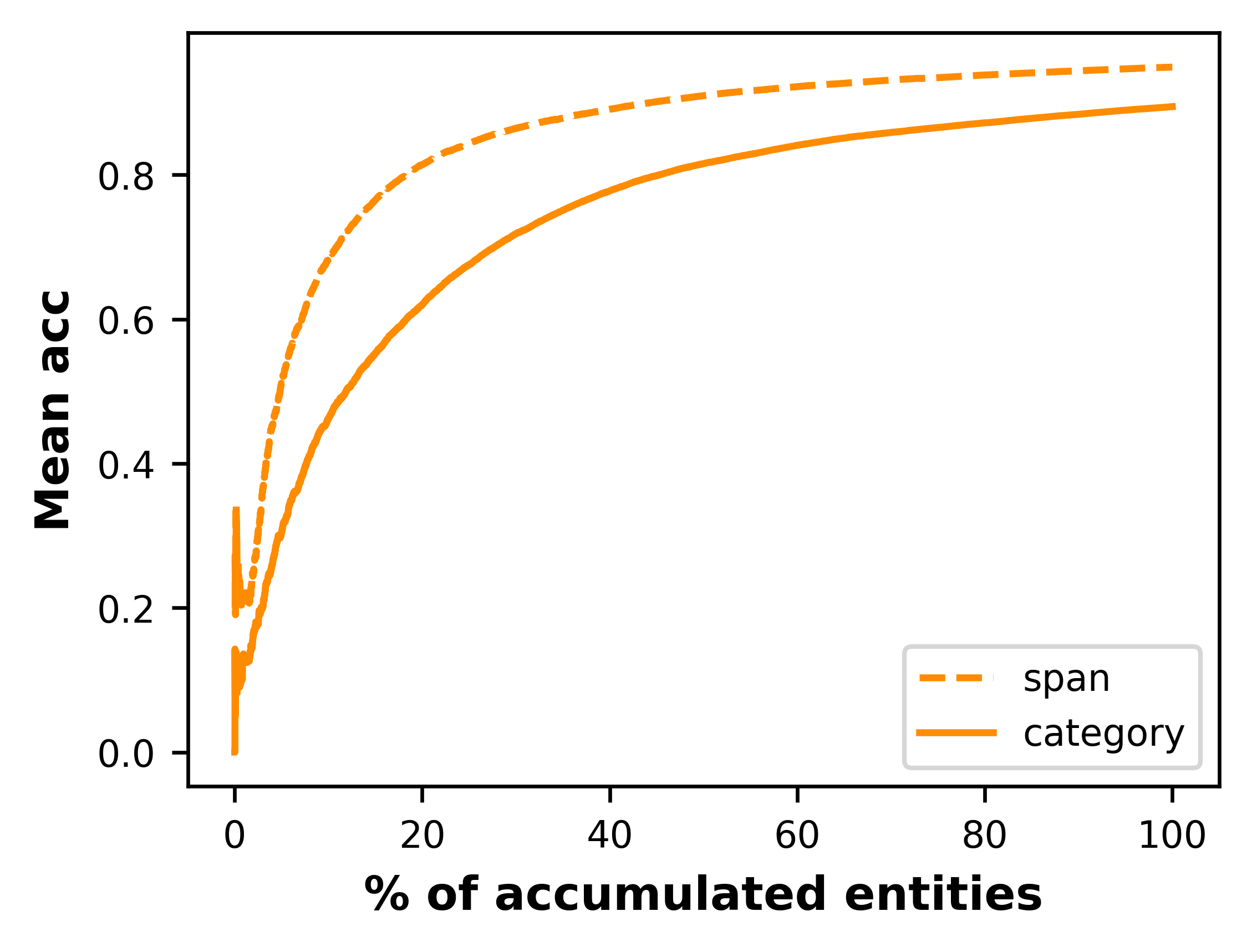}
    \caption{Mean accuracy of accumulated entities with ascending order of discriminator scores on CoNLL03 with $|C|=200$.} 
    \label{fig:dis_acc}
    \vspace{-3mm}
\end{figure}

We also investigate the quality of the discriminator in Figure \ref{fig:dis_acc}. We rank the pseudo entities in ascending order based on discriminator scores and then report the mean accuracy by comparing these pseudo entities with their corresponding ground-truth entities. As seen in the figure, entities with low discriminator scores exhibit poor quality. For instance, the accuracy of the category labels for the bottom 10\% of entities is approximately 0.4. As we incorporate more high-scoring entities, the mean accuracy shows a noticeable increase. This trend elucidates the discriminator's role in guiding the training of the NER model, primarily by accurately downweighting noisy labels.

\begin{table*}[htp!]
\centering
\caption{Case study of the discriminator. The sentences are from CoNLL03.}
\label{tab:case_dis}
\begin{tabular}{@{}l@{}}
\toprule\toprule

\textbf{Ground truth} \\ After the defeat of the resolution , drafted by the European Union and the United States , \textcolor{red}{[China]$_{\text{LOC}}$} \\ 's  Foreign Ministry thanked 26 countries for backing its motion for " no action " on the document .   \\ \midrule
After the defeat of the resolution , drafted by the European Union and the United States , China 's \\ Foreign Ministry thanked 26 countries for backing its motion for " no action " on the document . \\ \textcolor{magenta}{[China]} is a <mask> entity . \textcolor{green}{Span score is 0.9999}.            \\ \midrule
After the defeat of the resolution , drafted by the European Union and the United States , China 's \\ Foreign Ministry thanked 26 countries for backing its motion for " no action " on the document . \\ \textcolor{magenta}{[China]} is a <mask>  \textcolor{orange}{[organization]} entity . \textcolor{green}{Category score is 0.0003}. \\ \midrule \midrule
\textbf{Ground truth} \\
Arafat subsequently cancelled a meeting between Israeli and PLO officials , on civilian affairs , \\ at the Allenby Bridge crossing between Jordan and the \textcolor{red}{[West Bank]$_{\text{LOC}}$} . \\ \midrule
Arafat subsequently cancelled a meeting between Israeli and PLO officials , on civilian affairs , \\ at the Allenby Bridge crossing between Jordan and the West Bank . \textcolor{magenta}{[West Bank]} is a <mask> \\  entity . \textcolor{green}{Span score is 0.9993}. \\ \midrule
Arafat subsequently cancelled a meeting between Israeli and PLO officials , on civilian affairs , \\ at the Allenby Bridge crossing between Jordan and the West Bank . \textcolor{magenta}{[West Bank]} is a <mask> \\ \textcolor{orange}{[organization]} entity . \textcolor{green}{Category score is 0.0004}. \\ \midrule \midrule
\textbf{Ground truth} \\
At a news conference attended by approximately 50 players on Sunday , U.S. \textcolor{blue}{[Davis Cup]$_{\text{MISC}}$} \\ player Todd Martin expressed the players ' outrage at the seedings .\\ \midrule
At a news conference attended by approximately 50 players on Sunday , U.S. Davis Cup player \\ Todd Martin expressed the players ' outrage at the seedings . \textcolor{magenta}{[Davis]} is a <mask> entity . \\ \textcolor{green}{Span score is 0.0009}.\\ \midrule
At a news conference attended by approximately 50 players on Sunday , U.S. Davis Cup player \\ Todd Martin expressed the players ' outrage at the seedings . \textcolor{magenta}{[Davis]} is a <mask> \textcolor{blue}{[other]} entity . \\ \textcolor{green}{Category score is 0.0346}. \\ \bottomrule \bottomrule
\end{tabular}
\end{table*}

\end{document}